\documentclass{llncs}
\usepackage{graphicx}
\usepackage{subfigure}
\usepackage{url}
\usepackage{common}

\title{How Many Topics? \\Stability Analysis for Topic Models}
\author{Derek Greene, Derek O'Callaghan, P\'{a}draig Cunningham }
\institute{School of Computer Science \& Informatics, University College Dublin \\
\email{\{derek.greene,derek.ocallaghan,padraig.cunningham\}@ucd.ie} }

\begin{document} 
\maketitle
\begin{abstract}
Topic modeling refers to the task of discovering the underlying thematic structure in a text corpus, where the output is commonly presented as a report of the top terms appearing in each topic. Despite the diversity of topic modeling algorithms that have been proposed, a common challenge in successfully applying these techniques is the selection of an appropriate number of topics for a given corpus. Choosing too few topics will produce results that are overly broad, while choosing too many will result in the``over-clustering'' of a corpus into many small, highly-similar topics. In this paper, we propose a term-centric  stability analysis strategy to address this issue, the idea being that a model with an appropriate number of topics will be more robust to perturbations in the data. Using a topic modeling approach based on matrix factorization, evaluations performed on a range of corpora show that this strategy can successfully guide the model selection process. 

%
%



\end{abstract}


\section{Introduction}
\label{sec:intro}

From a general text mining perspective, a \emph{topic} in a text corpus can be viewed as either a probability distribution over the terms present in the corpus or a cluster that defines weights for those terms \cite{wang12group}. Considerable research on topic modeling has focused on the use of probabilistic methods such as variants of Latent Dirichlet Allocation (LDA) \cite{blei03lda} and Probabilistic Latent Semantic Analysis (PLSA) \cite{hofmann99probabilistic}. Non-probabilistic algorithms, such as Non-negative Matrix Factorization (NMF) \cite{lee99nmf}, have also been applied to this task \cite{wang12group,arora12beyond}. 
%
Regardless of the choice of algorithm, a key consideration in successfully applying topic modeling is the selection of an appropriate number of topics $k$ for the corpus under consideration. Choosing a value of $k$ that is too low will generate topics that are overly broad, while choosing a value that is too high will result in ``over-clustering'' of the data. For some corpora, coherent topics will exist at several different resolutions, from coarse to fine-grained, reflected by multiple appropriate $k$ values. 

When a clustering result is generated using an algorithm that contains a stochastic element or requires the selection of one or more key parameter values, it is important to consider whether the solution represents a ``definitive'' solution that may easily be replicated. Cluster validation techniques based on this concept have been shown to be effective in helping to choose a suitable number of clusters in data \cite{lange04article,levine01resampling}. The \emph{stability} of a clustering model refers to its ability to consistently replicate similar solutions on data originating from the same source. In practice, this involves repeatedly clustering using different initial conditions and/or applying the algorithm to different samples of the complete data set. A high level of agreement between the resulting clusterings indicates high stability, in turn suggesting that the current model is appropriate for the data. In contrast, a low level of agreement indicates that the model is a poor fit for the data. Stability analysis has most frequently been applied in bioinformatics \cite{brunet04metagenes,bertoni05stability}, where the focus has been on model selection for classical clustering approaches, such as $k$-means \cite{lange04article,bendavid07stability} and agglomerative hierarchical clustering \cite{levine01resampling,bertoni05stability}. 

In the literature, the output of topic modeling procedures is often presented in the form of lists of top-ranked terms suitable for human interpretation. Motivated by this,  we propose a term-centric stability approach for selecting the number of topics in a corpus, based on the agreement between term rankings generated over multiple runs of the same algorithm. We employ a ``top-weighted'' ranking measure, where higher-ranked terms have a greater degree of influence when calculating agreement scores. To ensure that a given model is robust against perturbations, we use both sampling of documents from a corpora and random matrix initialization to produce diverse collections of topics on which stability is calculated. 
Unlike previous applications of the concept of stability in NMF  \cite{brunet04metagenes} or LDA \cite{steyvers07topicmodels,dewall08stability}, our approach is generic in the sense that it does not rely on directly comparing probability distributions or topic-term matrices. So although we highlight the use of this method in conjunction with NMF, it could be applied in conjunction with other topic modeling and document clustering techniques. 

This paper is organized as follows. \refsec{sec:related} provides a brief overview of existing work in the areas of matrix factorization, stability analysis, and rank agreement. In \refsec{sec:methods} we discuss the problem of measuring the similarity between sets of term rankings, and describe a solution that can be used to quantify topic stability. Using a topic modeling approach based on matrix factorization, in \refsec{sec:eval} we present an empirical evaluation of the proposed solution on a range of text corpora. The paper finishes with some conclusions and suggestions for future work in \refsec{sec:conc}.
\section{Related Work}
\label{sec:related}

\subsection{Matrix Factorization}

While work on topic models has largely focused on the use of LDA \cite{blei03lda,steyvers07topicmodels}, Non-negative Matrix Factorization (NMF)  can also be applied to textual data to reveal topical structures \cite{wang12group}. NMF seeks to decompose a data matrix into factors that are constrained so that they will not contain negative values. 
%
%
Given a document-term matrix $\m{A} \in \Real^{m\times n}$ representing $m$ unique terms present in a corpus of $n$ documents, NMF generates a reduced rank-$k$ approximation in the form of the product of two non-negative factors $\m{A} \approx \m{W}\m{H}$, where the objective is to minimize the reconstruction error between $\m{A}$ and the low-dimensional approximation. The columns or \emph{basis vectors} of $\m{W} \in \Real^{m\times k}$ can be interpreted as topics, defined with non-negative weights relative to the $m$ terms. The entries in the matrix $\m{H} \in \Real^{k\times n}$ provide document memberships with respect to the $k$ topics.
Note that, unlike LDA which operates on raw frequency counts, NMF can be applied to a non-negative matrix $\m{A}$ that has been previously normalized using common pre-processing procedures such as TF-IDF term weighting and document length normalization. As with LDA, document-topic assignments are not discrete, allowing a single document to be associated with multiple topics.

For NMF, the key model selection challenge is the selection of the user-defined parameter $k$. Although no definitive approach for choosing $k$ has been identified, a number of heuristics exist in the literature. A simple technique is to calculate the Residual Sum of Squares (RSS) between the approximation given by a pair of NMF factors and the original matrix \cite{hutchins08position}, which indicates the degree of variation in the dependent variables the NMF model did not explain.  The authors suggest that, by examining the RSS curve for a range of candidate values of $k$, an inflection point might be identified to provide a robust estimate of the optimal reduced rank. 

\subsection{Stability Analysis}

A range of methods based on the concept of \emph{stability analysis} have been proposed for the task of model selection. The \emph{stability} of a clustering algorithm refers to its ability to consistently produce similar solutions on data originating from the same source \cite{lange04article,bendavid07stability}. Since only a single set of data items will be generally available in unsupervised learning tasks, clusterings are generated on perturbations of the original data. The primary application of stability analysis has been as a robust approach for selecting key algorithm parameters \cite{law03boot}, specifically when estimating the optimal number of clusters for a given data set. These methods are motivated by the observation that, if the number of clusters in a model is too large, repeated clusterings will lead to arbitrary partitions of the data, resulting in unstable solutions. On the other hand, if the number of clusters is too small, the clustering algorithm will be constrained to merge subsets of objects which should remain separated, also leading to unstable solutions. In contrast, repeated clusterings generated using some optimal number of clusters  will generally be consistent, even when the data is perturbed or distorted.

The most common approach to stability analysis involves perturbing the data by randomly sampling the original objects to produce a collection of subsamples for clustering using values of $k$ from a pre-defined range \cite{levine01resampling}. The stability of the clustering model for each candidate value of $k$ is evaluated using an agreement measure evaluated on all pairs of clusterings generated on different subsamples. One or more values of $k$ are then recommended, selected based on the highest mean agreement scores.
%

Brunet~\etal proposed an initial stability-based approach for  NMF model selection based on discretized cluster assignments of items (rather than features) across multiple runs of the same algorithm using different random initializations \cite{brunet04metagenes}. Specifically, for each NMF run applied to the same data set of $n$ items, a $n \times n$ \emph{connectivity matrix} is constructed, where an entry $(i,j) = 1$ if items $i$ and $j$ are assigned to the same discrete cluster, and $(i,j) = 0$ otherwise.  By repeating this process over $\tau$ runs, a \emph{consensus matrix} can be calculated as the average of all $\tau$ connectivity matrices. Each entry in this matrix indicates the fraction of times two items were clustered together. To measure the stability of a particular value of $k$, a cophenetic correlation coefficient is calculated on a hierarchical clustering of the connectivity matrix. The authors suggest a heuristic for selecting one or more values of $k$, based on a sudden drop in the correlation score as $k$ increases. 

In their work on LDA, Steyvers and Griffiths noted the importance of identifying those topics that will appear repeatedly across multiple samples of related data \cite{steyvers07topicmodels}, which closely resembles the more general concept of stability analysis \cite{levine01resampling}. The authors suggested comparing two runs of LDA by examining a topic-topic matrix constructed from the symmetric Kullback Liebler (KL) distance between topic distributions from the two runs.
Alternative work on measuring the stability of LDA topic models was described in \cite{dewall08stability}. The authors proposed a document-centric approach, where topics from two different LDA runs are matched together based on correlations between rows of the two corresponding document-topic matrices. The output was represented as a document-document correlation matrix, where block diagonal structured induced by the correlation values are indicative of higher stability. In this respect, the approach is similar to the Brunet~\etal approach for NMF. 

Other evaluation measures used for LDA have included those based on the semantic coherence of the top terms derived from a single set of topics, with respect to term co-occurrence within the same corpus or an external background corpus. For example, Newman \etal
calculated correlations between human judgements and a set of proposed measures, and found that a Pointwise Mutual Information (PMI)
measure achieved best or near-best out of all those considered \cite{newman10aet}. However, such measures have not focused on model selection and do not consider the robustness of topics over multiple runs of an algorithm.

\subsection{Ranking Comparison}

A variety of well-known simple metrics exist for measuring the distance or similarity between pairs of ranked lists of the same set of items, notably Spearman's footrule distance and Kendall's tau function \cite{kendall90rank}. 
However, Webber~\etal \cite{webber10indefinite} note that many problems will involve comparing \emph{indefinite rankings}, where items appear in one list but not in another list, but standard metrics do not consider such cases.
For other applications, it will be desirable to employ a \emph{top-weighted} ranking agreement measure, such that changing the rank of a highly-relevant item at the top of a list results in a higher penalty than changing the rank of an irrelevant item appearing at the tail of a list. This consideration is important in the case of comparing query results from different search engines, though, as we demonstrate later, it is also a key consideration when comparing rankings of terms arising in topic modeling. 

Motivated by basic set overlap, Fagin~\etal \cite{fagin03topk} proposed a top-weighted distance metric between indefinite rankings, also referred to as Average Overlap (AO) \cite{webber10indefinite}, which calculates the mean intersection size between every pair of subsets of $d$ top-ranked items in two lists, for $d=[1,t]$. This naturally accords a higher positional weight to items at the top of the lists.
More recently, Kumar and Vassilvitskii proposed a generic framework for measuring the distance between a pair of rankings \cite{kumar10rankings}, supporting both positional weights and item relevance weights. Based on this framework, generalized versions of Kendall's tau and Spearman's footrule metric were derived. However, the authors did not focus on the case of indefinite rankings.

\section{Methods}
\label{sec:methods}

In this section we describe a general stability-based method for selecting the number of topics for topic modeling. Unlike previous unsupervised stability analysis methods, we focus on the use of features or terms to evaluate the suitability of a model. This is motivated by the term-centric approach generally taken in topic modeling, where precedence is generally given to the term-topic output and topics are summarized using a truncated set of top terms. Also, unlike the approach proposed in \cite{brunet04metagenes} for genetic data, our method does not assume that topic clusters are entirely disjoint and does not require the calculation of a dense connectivity matrix or the application of a subsequent clustering algorithm.

Firstly, in \refsec{sec:termsim} we describe a similarity metric for comparing  two ranked lists of terms. Using this measure, in \refsec{sec:stability} we propose a measure of the agreement between two topic models when represented as ranked term lists. Subsequently, in \refsec{sec:select} we propose a stability analysis method for selecting the number of topics in a text corpus.


\subsection{Term Ranking Similarity}
\label{sec:termsim}

A general way to represent the output of a topic modeling algorithm is in the form of a \emph{ranking set} containing $k$ ranked lists, denoted $\aset{S}=\fullset{R}{k}$. The $i$-th topic produced by the algorithm is represented by the list $R_{i}$, containing the top $t$ terms which are most characteristic of that topic according to some criterion. In the case of NMF, this will correspond to the highest ranked values in each column of the $k$ basis vectors, while for LDA this will consist of the terms with the highest probabilities in the term distribution for each topic. For partitional or hierarchical document clustering algorithms, this might consist of the highest ranked terms in each cluster centroid.

A variety of symmetric measures could be used to assess the similarity between a pair of ranked lists $(R_{i},R_{j})$. A \naive approach would be to employ a simple set overlap method, such as the Jaccard index \cite{jaccard12index}. However, such measures do not take into account positional information. Terms occurring at the top of a ranked list generated by an algorithm such as NMF will naturally be more relevant to a topic than those occurring at the tail of the list, which correspond to zero or near-zero values in the original basis vectors. Also, in practice, rather than considering all $m$ terms in a corpus, the results of topic modeling are presented using the top $t << m$ terms. Similarly, when measuring the similarity between ranked lists, it may be preferable to consider truncated lists with only $t$ terms, for economy of representation and to reduce the computational cost of applying multiple similarity operations. However, this will often lead to indefinite rankings, where different subsets of terms are being compared.

Therefore, following the ranking distance measure proposed by Fagin~\etal \cite{fagin03topk}, we propose the use of a top-weighted version of the Jaccard index, suitable for calculating the similarity between pairs of indefinite rankings. Specifically, we define the \emph{Average Jaccard} (AJ) measure as follows. We calculate the average of the Jaccard scores between every pair of subsets of $d$ top-ranked terms in two lists, for depth $d \in [1,t]$. That is:
\begin{equation}
AJ(R_{i},R_{j})=\frac{1}{t} \sum_{d=1}^{t} \gamma_{d}(R_{i},R_{j})
\label{eqn:aj}
\end{equation}
where 
\begin{equation}
\gamma_{d}(R_{i},R_{j}) = \frac{\abs{ R_{i,d} \cap R_{j,d} } }{ \abs{ R_{i,d} \cup R_{j,d} } }
\end{equation}
such that $R_{i,d}$ is the head of list $R_{i}$ up to depth $d$.
This is a symmetric measure producing values in the range $[0,1]$, where the terms through a ranked list are weighted according to a decreasing linear scale. To demonstrate this, a simple illustrative example is shown in \reftab{tab:aj}. Note that, although the Jaccard score at depth $d=5$ is comparatively high (0.429), the mean score is much lower (0.154), as the similarity between terms occurs towards the tails of the lists -- these terms carry less weight than those at the head of the lists, such as ``album'' and ``sport''. 

\begin{table}[!t]
\centering
\caption{Example of Average Jaccard (AJ) term ranking similarity, for two ranked lists of terms up to depth $d=5$. The value $\textrm{Jac}_{d}$ indicates the Jaccard score at depth $d$ only, while ${AJ}$ indicates the current AJ similarity at that depth.}
\label{tab:aj}
\begin{tabular}{|c|p{4.5cm}|p{4.5cm}|c|c|}
\hline
$\;d\;$ & $R_{1,d}$                                & $R_{2,d}$                                & $\textrm{Jac}_{d}$  & $AJ$    \\
\hline
1     & album                          & sport                          & \;\;0.000\;\; & \;\;0.000\;\; \\
2     & album, music                   & sport, best                    & 0.000 & 0.000 \\
3     & album, music, best             & sport, best, win               & 0.200 & 0.067 \\
4     & album, music, best, award      & sport, best, win, medal        & 0.143 & 0.086 \\
5     & album, music, best, award, win & sport, best, win, medal, award & 0.429 & 0.154 \\
\hline
\end{tabular}
\end{table}

\subsection{Topic Model Agreement}
\label{sec:stability}

\begin{figure}[!t]
    \centering
    \includegraphics[width=0.95\linewidth]{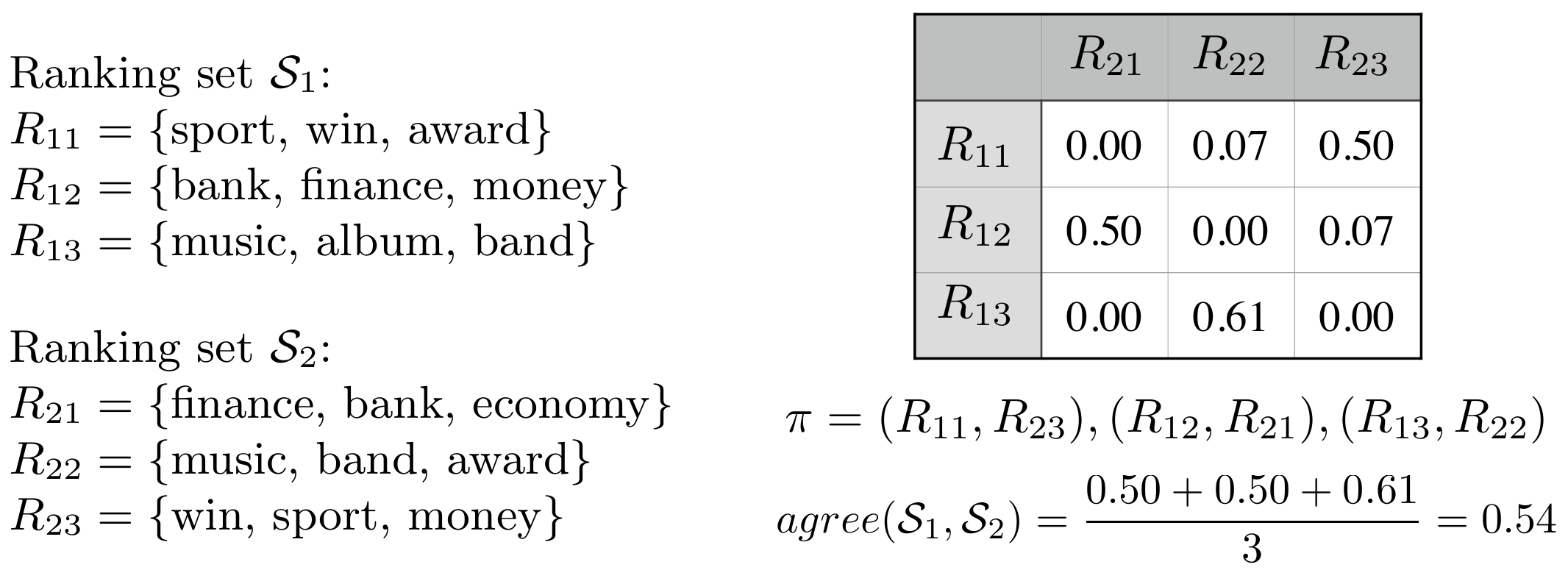}
	\vskip -0.6em
	\caption{A simple example of measuring the agreement between two different topic models, each containing $k=3$ topics, represented by a pair of ranking sets. Term ranking similarity values are calculated using Average Jaccard, up to depth $d=3$.}
    \label{fig:example}
\end{figure}

We now consider the problem of measuring the agreement between two different $k$-way topic models, represented as two ranking sets $\aset{S}_{x}=\{R_{x1},\dots,R_{xk}\}$ and $\aset{S}_{y}=\{R_{y1},\dots,R_{yk}\}$, both containing $k$ ranked lists. We construct a $k \times k$ similarity matrix $\m{M}$, such that the entry $M_{ij}$ indicates the agreement between $R_{xi}$ and $R_{yj}$ (\ie the $i$-th topic in the first model and the $j$-th topic in the second model), as calculated using the Average Jaccard score (\reft{eqn:aj}). 
We then find the best match between the rows and columns of $\m{M}$ (\ie the ranked lists in $\aset{S}_{x}$ and $\aset{S}_{y}$). The optimal permutation $\pi$ may be found in $O(k^{3})$ time by solving the minimal weight bipartite matching problem using the Hungarian method \cite{kuhn55hungarian}. From this, we can produce an agreement score:
\begin{equation}
agree(\aset{S}_{x},\aset{S}_{y}) = \frac{1}{k}\sum_{i=1}^{k} AJ(R_{xi},\pi(R_{xi}))
\label{eqn:agree}
\end{equation}
where $\pi(R_{xi})$ denotes the ranked list in $\aset{S}_{y}$ matched to $R_{xi}$ by the permutation $\pi$. Values for the above take the range $[0,1]$, where a comparison between two identical $k$-way topic models will result in a score of 1. A simple example illustrating the agreement process is shown in \reffig{fig:example}.

\subsection{Selecting the Number of Topics}
\label{sec:select}

Building on the agreement measure defined in \refsec{sec:stability}, we now propose a model selection approach for topic modeling. For each value of $k$ in a broad pre-defined range $[k_{min},k_{max}]$, we proceed as follows. We firstly generate an initial topic model on the complete data set using an appropriate  algorithm (ideally this should be deterministic in nature), which provides a reference point for analyzing the stability afforded by using $k$ topics. We represent this as a \emph{reference ranking set} $\aset{S}_{0}$, where each topic is represented by the ranked list of its top $t$ terms.
Subsequently, $\tau$ samples of the data set are constructed by randomly selecting a subset of $\beta \times n$ documents without replacement, where $0 \leq \beta \leq 1$ denotes the sampling ratio controlling the number of documents in each sample. We then generate $\tau$ $k$-way topic models by applying the topic modeling algorithm to each of the samples, resulting in alternative ranking sets $\{ \aset{S}_{1}, \dots, \aset{S}_{\tau} \}$, where all topics are also represented using $t$ top terms. To measure the overall stability at $k$, we calculate the mean agreement between the reference ranking set and all other ranking sets using \reft{eqn:agree}:
\begin{equation}
stability(k) = \frac{1}{\tau}\sum_{i=1}^{\tau} agree(\aset{S}_{0}, \aset{S}_{i})
\label{eqn:stability}
\end{equation}
This process is repeated for each candidate $k \in [k_{min},k_{max}]$.  A summary of the entire process is given in \reffig{fig:alg}. Note that the proposed approach is similar to the strategy for item stability analysis proposed in \cite{levine01resampling}, in that a single reference point is used for each value of $k$, involving $\tau$ comparisons between solutions. This contrasts with the approach used by other authors in the literature (\eg \cite{law03boot}) which involves comparing all unique pairs of results, requiring $\frac{\tau \times (\tau-1)}{2}$ agreement comparisons. 

\begin{figure}[!t]
\begin{algor}
\item Randomly generate $\tau$ samples of the data set, each containing $\beta \times n$ documents.
\item For each value of $k \in [k_{min},k_{max}]$ :
	\begin{enumerate}
	\item Apply the topic modeling algorithm to the complete data set of $n$ documents to generate $k$ topics, and represent the output as the reference ranking set $\aset{S}_{0}$.
	\item For each sample $\m{X}_{i}$:
		\begin{enumerate}
		\item Apply the topic modeling algorithm to $\m{X}_{i}$ to generate $k$ topics, and represent the output as the ranking set $\aset{S}_{i}$.
		\item Calculate the agreement score $agree(\aset{S}_{0},\aset{S}_{i})$.
		\end{enumerate}
	\item Compute the mean agreement score for $k$ over all $\tau$ samples (\reft{eqn:stability}).
	\end{enumerate}
\item Select one or more values for $k$ based upon the highest mean agreement scores.	
\end{algor}
\caption{Summary of the proposed stability analysis method for topic models.}
\label{fig:alg}
\end{figure}

\begin{figure}[!t]
\center
\subfigure[]{\includegraphics[width=0.48\linewidth]{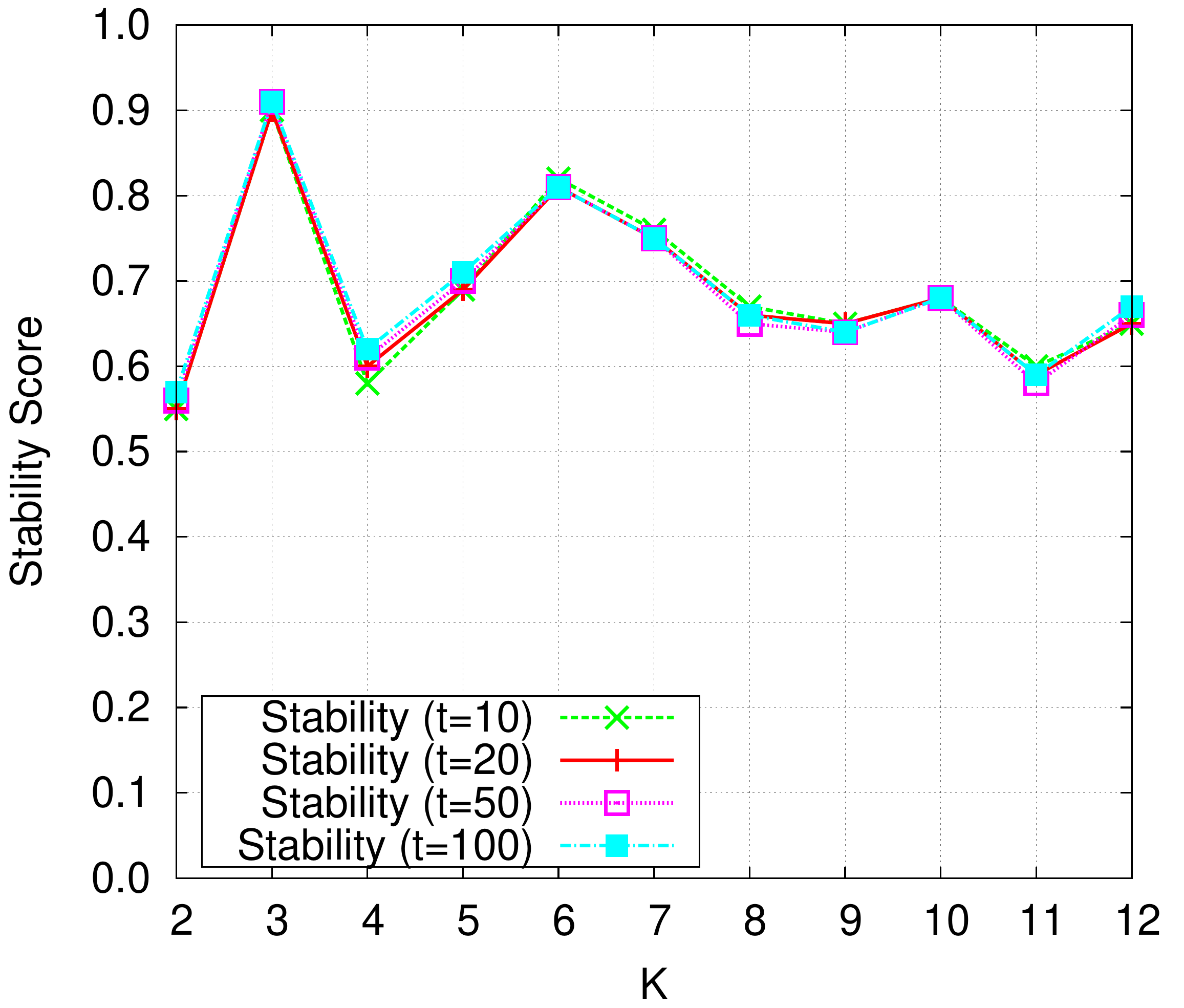}}
\hskip 0.8em
\subfigure[]{\includegraphics[width=0.48\linewidth]{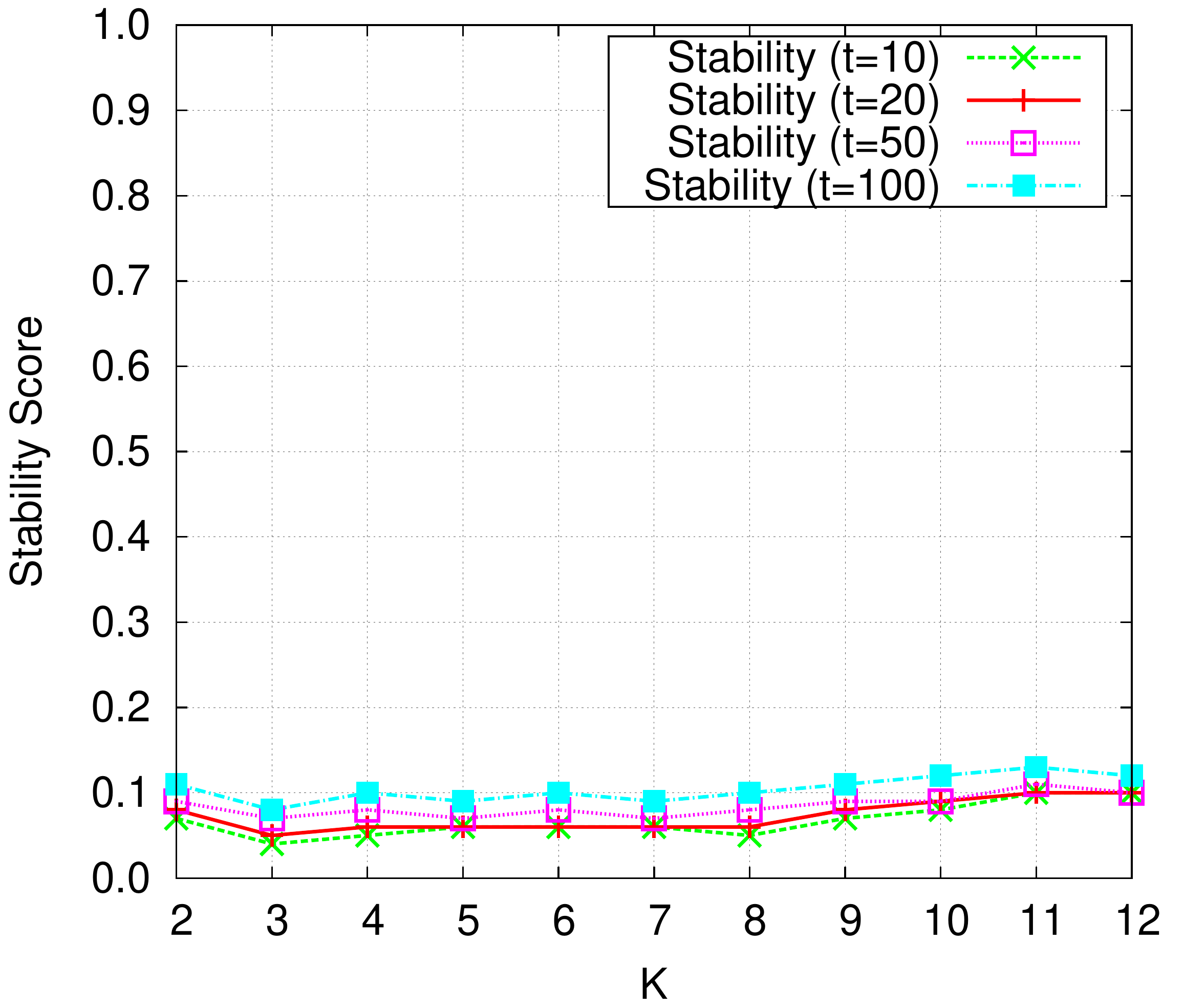}}
\vskip -0.4em
\caption{Stability analysis plots generated using $t=10/20/50/100$ top terms for (a) the \emph{guardian-2013} corpus of news articles, (b) a synthetic dataset of 500 documents generated randomly from 1,500 terms.}
\label{fig:egplots}
\end{figure}

By examining a plot of the stability scores produced with \reft{eqn:stability}, a final value $k$ may be identified based on peaks in the plot. The presence of more than one peak indicates that multiple appropriate topic schemes exist for the corpus under consideration, which is analogous to the  existence of multiple alternative solutions in many general cluster analysis problems \cite{bae06coala}. An example of this case is shown in \reffig{fig:egplots}(a) for the \emph{guardian-2013} corpus. This data set has six annotated category labels, but we also see a peak at $k=3$ in the stability plots, suggesting that thematic structure exists at a more coarse level too.  On the other hand, a flat curve with no peaks, combined with low stability values, strongly suggests that no coherent topics exist in the data set. This is analogous to the general problem of identifying ``clustering tendency'' \cite{levine01resampling}. The example in \reffig{fig:egplots}(b) shows plots generated for a synthetic data set of 500 randomly generated documents. As one might expect, no strong peak appears in the stability plots.

\section{Evaluation}
\label{sec:eval}

\subsection{Data}

We now evaluate the stability analysis method proposed in \refsec{sec:methods} to assess its usefulness in guiding the selection of the number of topics for NMF. The evaluation is performed on a number of text corpora, each of which has annotated ``ground truth'' document labels, such that each document is assigned a single label. 
When pre-processing the data, terms occurring in $< 20$ documents were removed, along with English language stop words, but no stemming was performed. Standard log TF-IDF and L2 document length normalization procedures were then applied to the term-document matrix. Descriptions of the corpora are provided in \reftab{tab:data}, and pre-processed versions are made available online for further research\footnote{\url{http://mlg.ucd.ie/howmanytopics/}}.

\subsection{Experimental Setup}

In our experiments we compare the proposed stability analysis method with a popular existing approach for selecting the reduced rank for NMF based on the cophenetic correlation of a consensus matrix \cite{brunet04metagenes}. The experimental process involved applying both schemes to each corpus across a reasonable range of values for $k$, and comparing plots of their output. Here we use $k \in [2,12]$, based on the fact that the numbers of ground truth labels in the corpora listed in \reftab{tab:data} are within this range.

To provide a fair comparison, both schemes use information coming from the same collection of matrix factorizations. These were generated using the fast alternating least squares variant of NMF introduced by \cite{lin07gradient}, with random initialization to samples of the data. In all cases we allowed the factorization process to run for a maximum of 50 iterations. We use a sampling ratio of $\beta=0.8$ (\ie 80\% of documents are randomly chosen for each run), with a total of $\tau=100$ runs to minimize any variance introduced by sampling. 
For our stability analysis method, we also generate reference ranking sets for each candidate value of $k$ by applying NMF to the complete data set with Nonnegative Double Singular Value Decomposition (NNDSVD) initialization to ensure a deterministic solution \cite{bout08headstart}.

\begin{table}[!t]
	\centering
    \caption{Details of the corpora used in our experiments, including the total number of documents $n$, terms $m$, and number of labels $\hat{k}$ in the associated ``ground truth''.}
    \label{tab:data}
    \begin{tabular}{|p{2.3cm}|r|r|r|p{6.9cm}|}
    \hline
    \emph{Corpus}             & $n$\; & $m$\; & $\hat{k}$\;\ & \emph{Description} \\
    \hline \hline
    bbc                 & 2,225\;     & 3,121\;     & 5\;  & General news articles from the BBC \cite{greene05pkdd}. \\
   bbc-sport            & 737\;      & 969\;      & 5\; & Sports news articles from the BBC \cite{greene05pkdd}. \\
    guardian-2013   	& 6,520\;     & 10,801\;    & 6\;  &  New corpus of news articles published by The Guardian during 2013. \\
   irishtimes-2013    & 3,246\;     & 4,832\;     & 7\;  & New corpus of news articles published by The Irish Times during 2013.\\
	nytimes-1999            & 9,551\;     & 12,987\;   & 4\; & A subset of the New York Times Annotated Corpus from 1999 \cite{sandhaus08nytimes}. \\
    nytimes-2003            & 11,527\;     & 15,001\;   & 7\; & As above, with articles from 2003.  \\
    wikipedia-high      & 5,738\;     & 17,311\;    & 6\; & Subset of a Wikipedia dump from January 2014, where articles are assigned labels based on their high level WikiProject.  \\
   wikipedia-low       & 4,986\;     & 15,441\;    & 10\; & Another Wikipedia subset. Articles are labeled with fine-grained WikiProject sub-groups. \\ 
    \hline
    \end{tabular}
\end{table}

\subsection{Model Selection}
Initially, for stability analysis we examined a range of values $t=10/20/50/100$ for the number of top terms used to represent each topic when measuring agreement between ranked lists. However, the resulting stability scores generated for each value of $t$ were highly correlated across all corpora considered in our evaluation (see \reftab{tab:correl} for average correlations). A typical example of this behavior is shown in \reffig{fig:egplots}(a) for the \emph{guardian-2013} corpus, where the plots almost perfectly overlap. This behavior is perhaps unsurprising as, given the definition of the Average Jaccard measure in \reft{eqn:aj}, terms occurring further down ranked lists will naturally carry less weight. Therefore, the difference between scores generated with, say $t=50$ and $t=100$ will be minimal. For the remainder of this section we report stability scores for $t=20$, which provided the highest pairwise mean correlation (0.977) with the results from other values of $t$ examined, while also providing economy of representation for topics.

\begin{table}[!b]
\centering
\caption{Pearson correlation coefficient scores between stability scores for different numbers of top terms $t$, as averaged across all corpora in our evaluations.}
\label{tab:correl}
\begin{tabular}{|l|c|c|c|c||c|}
\hline
\emph{\;\# Terms}\;\;\;\;& \;$t=10$\;    & \;$t=20$\; & \;$t=50$\; & \;$t=100$\; & \;\emph{Mean}\; \\ \hline
\;$t=10$   & - & 0.964 & 0.929 & 0.926 & 0.940 \\ \hline
\;$t=20$  & 0.964 & - & 0.985 & 0.982 & {\bf{0.977}} \\ \hline
\;$t=50$  & 0.929 & 0.985 & - & 0.997 & 0.970 \\ \hline

\;$t=100$  & 0.926 & 0.982 & 0.997 & - & 0.968 \\ \hline
\end{tabular}
\end{table}

\begin{figure}[!t]
\center
\subfigure[\emph{bbc}]{\includegraphics[width=0.48\linewidth]{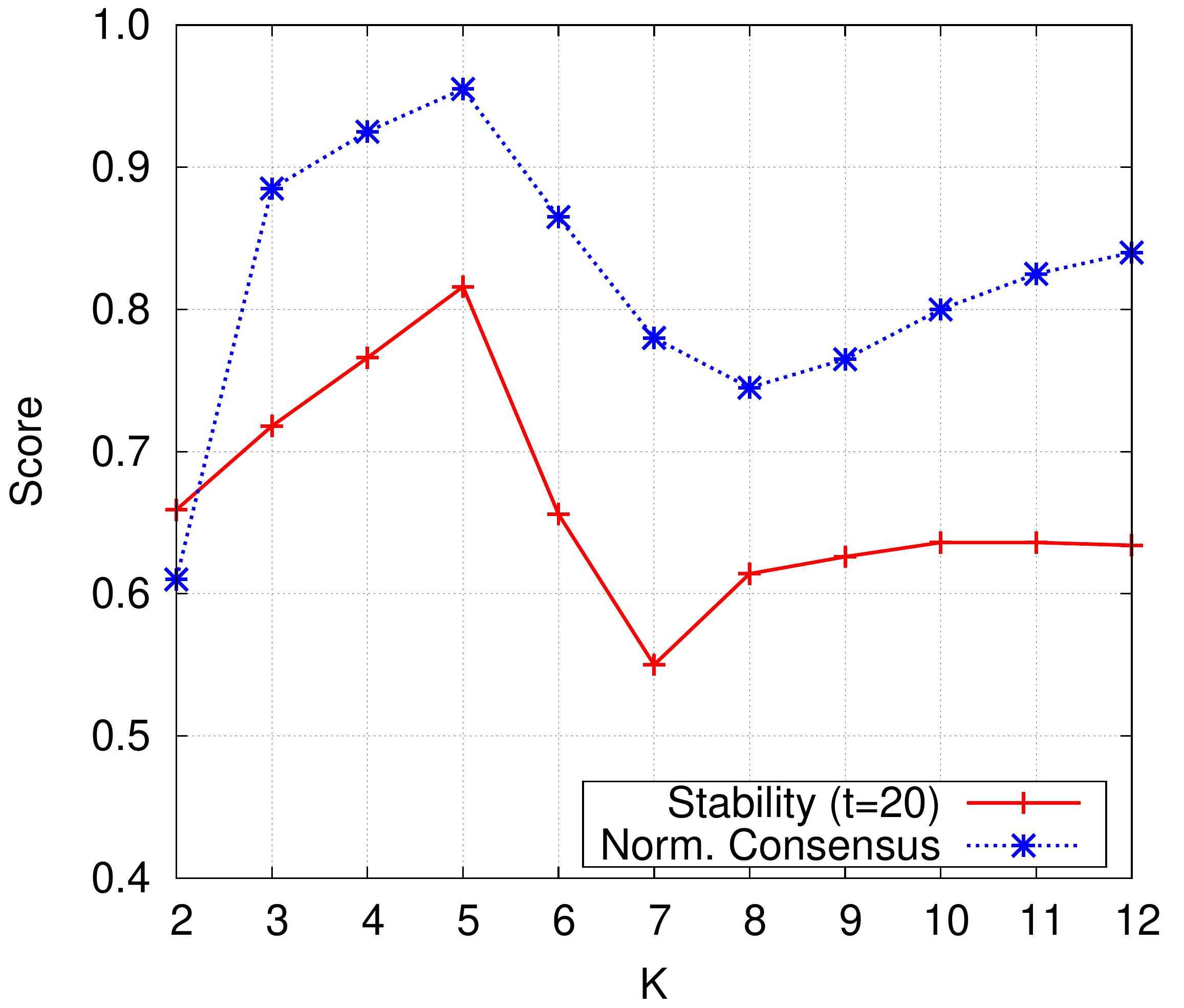}}
\hskip 0.8em
\subfigure[\emph{bbc-sport}]{\includegraphics[width=0.48\linewidth]{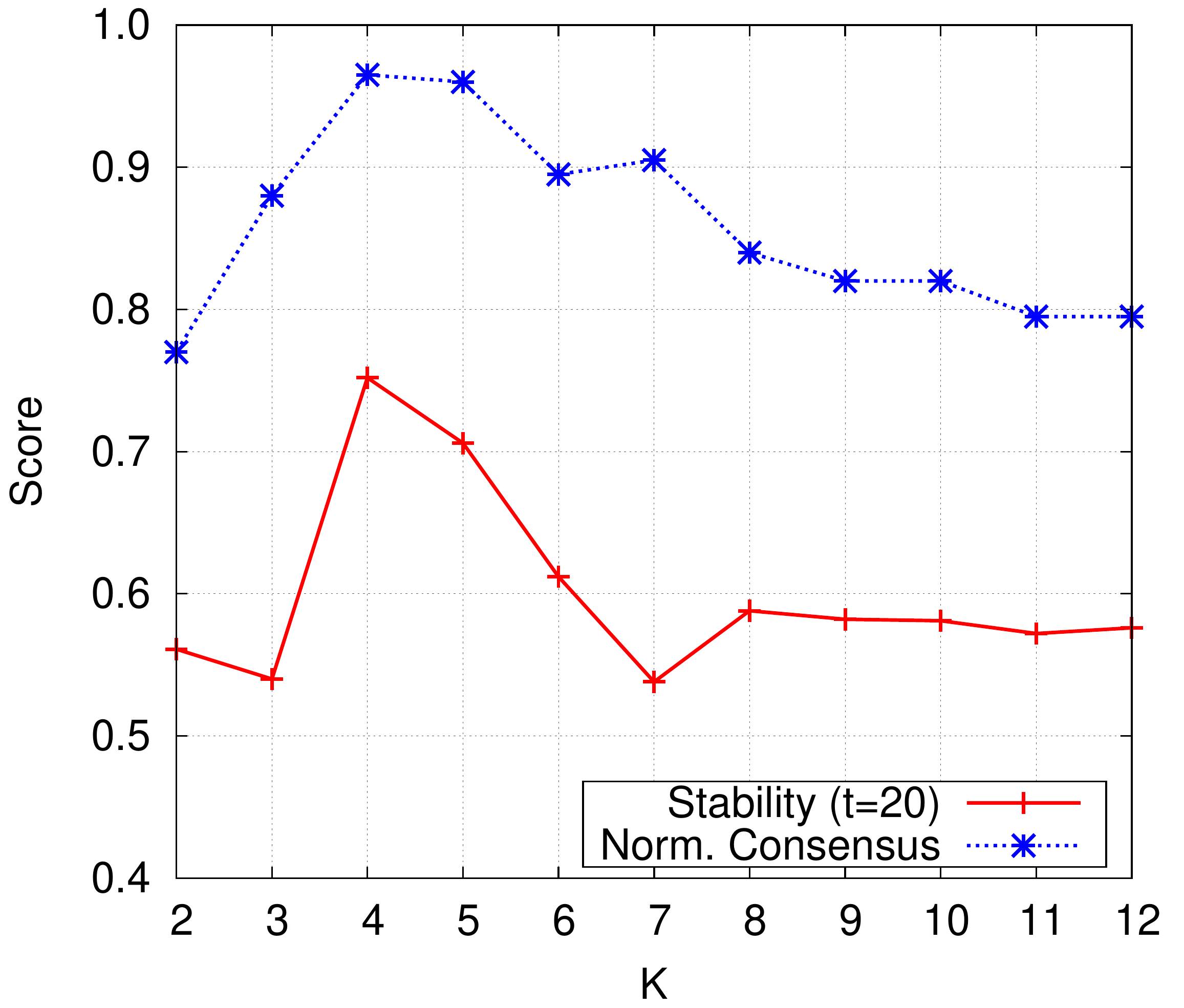}}
\subfigure[\emph{guardian-2013}]{\includegraphics[width=0.48\linewidth]{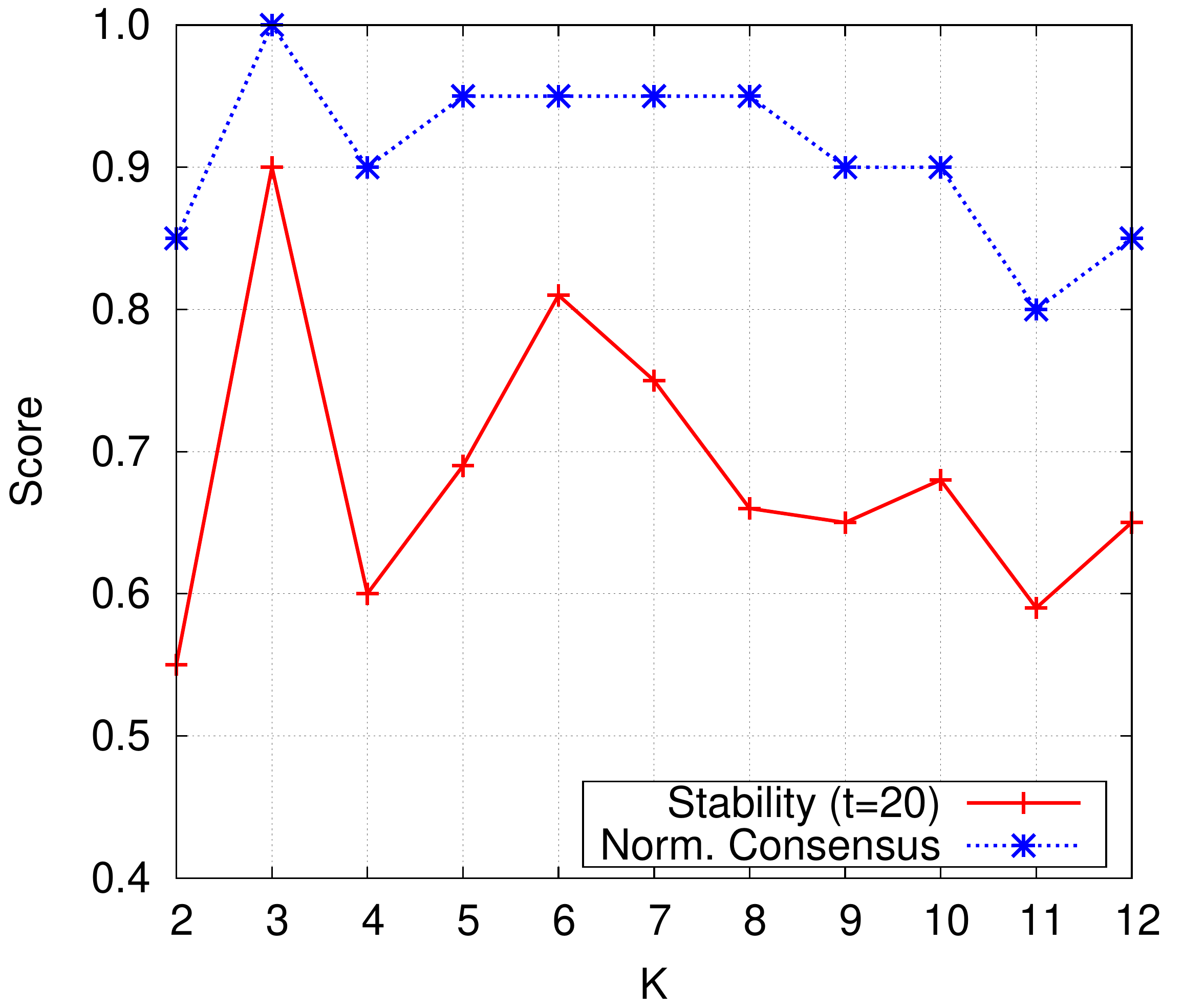}}
\hskip 0.8em
\subfigure[\emph{irishtimes-2013}]{\includegraphics[width=0.48\linewidth]{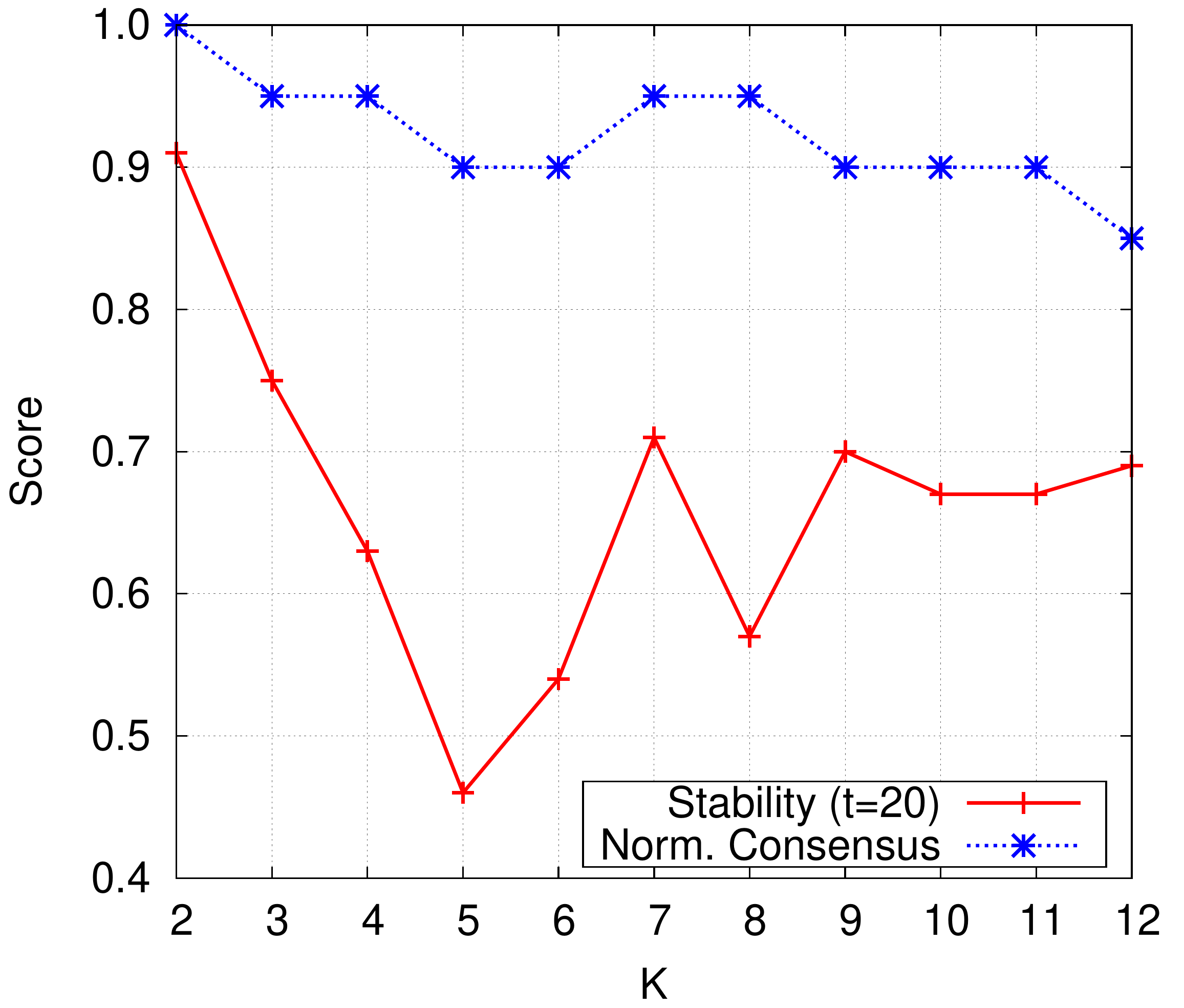}}
\vskip -1em
\caption{Comparison of plots generated for stability analysis ($t=20$) and consensus matrix analysis for values of $k \in [2,12]$. In both cases we attempt to identify one or more suitable values for $k$ based on peaks in the plots.}
\label{fig:results1}
\end{figure}

\begin{figure}[!t]
\center
\subfigure[\emph{nytimes-1999}]{\includegraphics[width=0.48\linewidth]{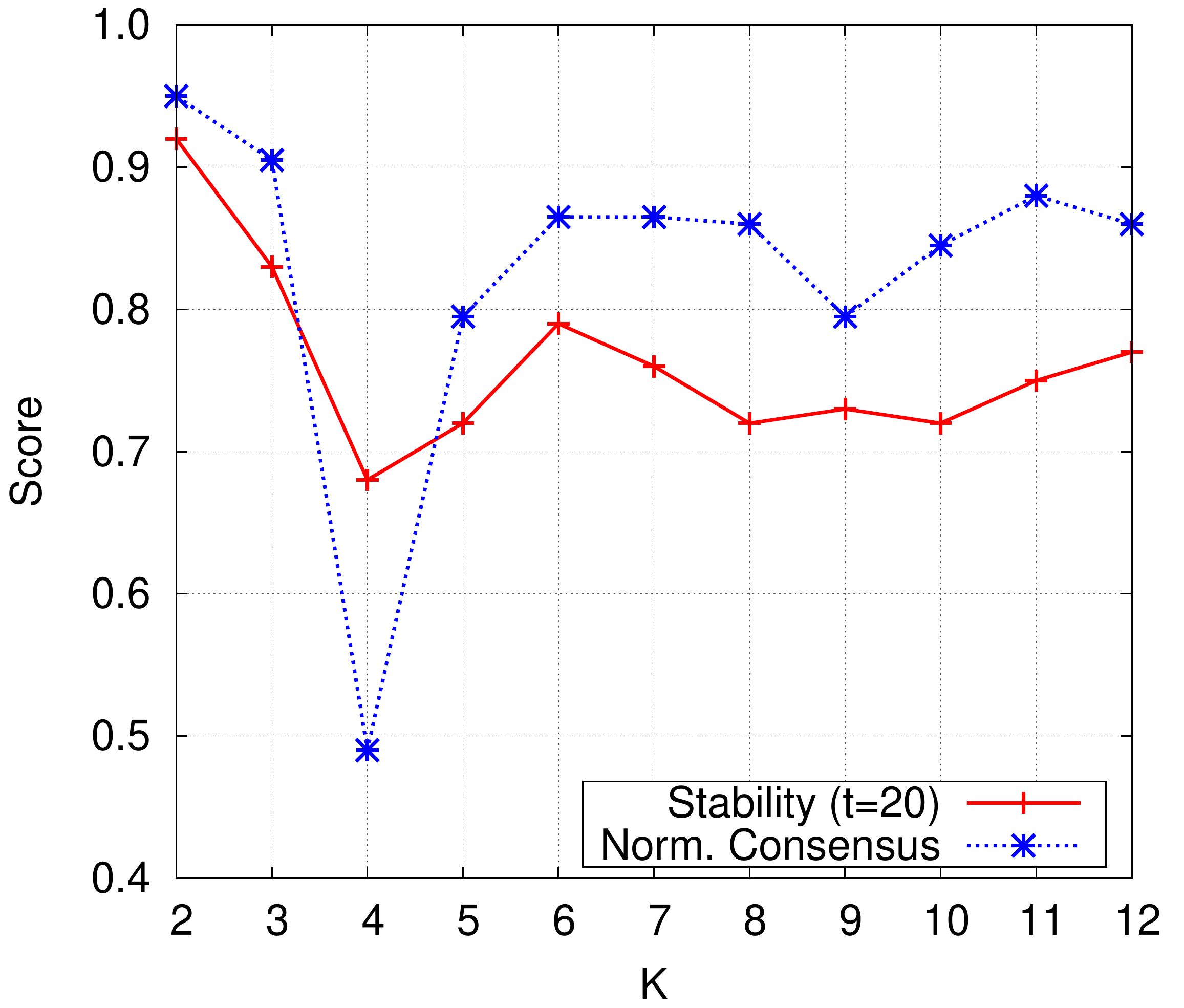}}
\hskip 0.8em
\subfigure[\emph{nytimes-2003}]{\includegraphics[width=0.48\linewidth]{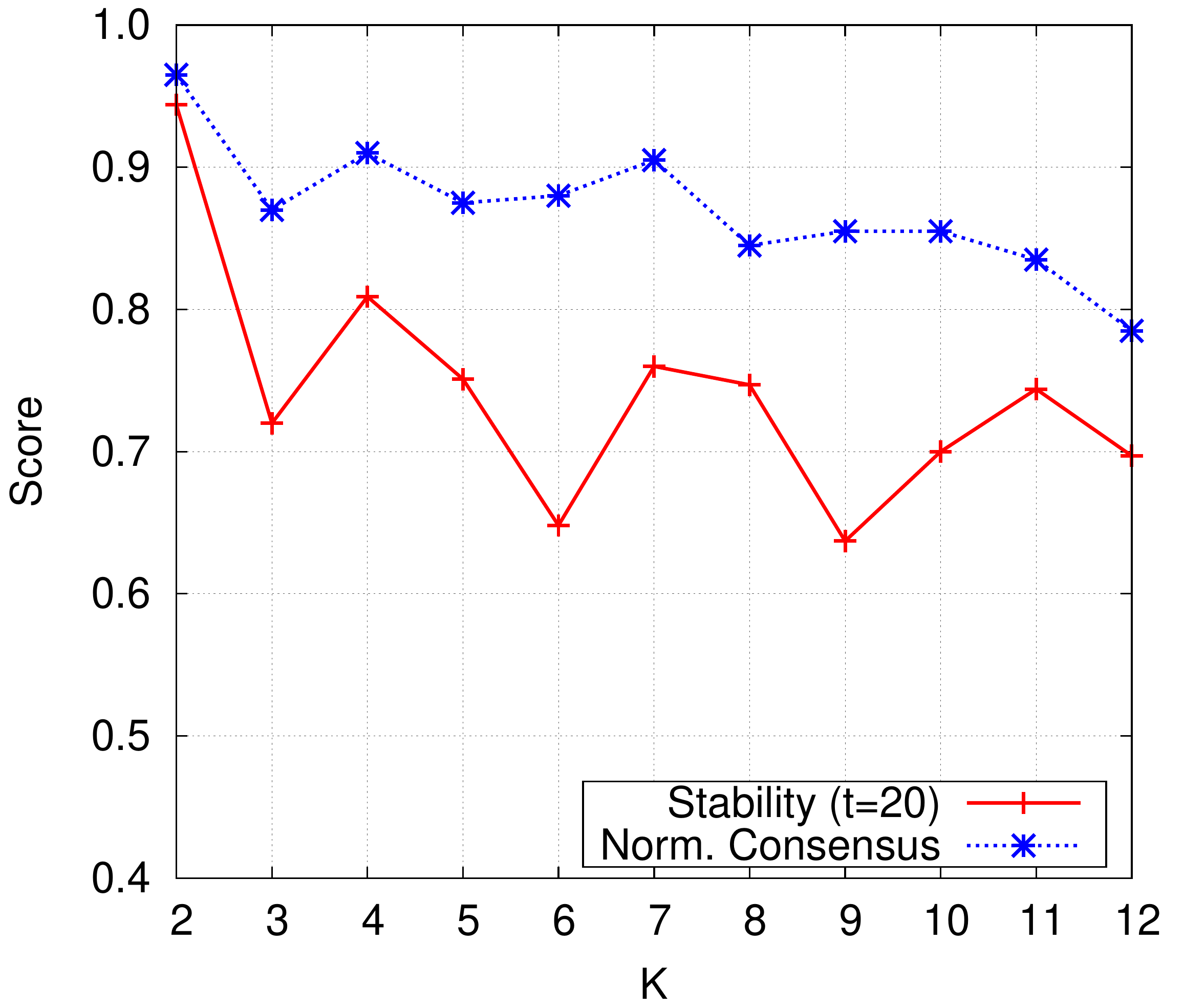}}
\subfigure[\emph{wikipedia-high}]{\includegraphics[width=0.48\linewidth]{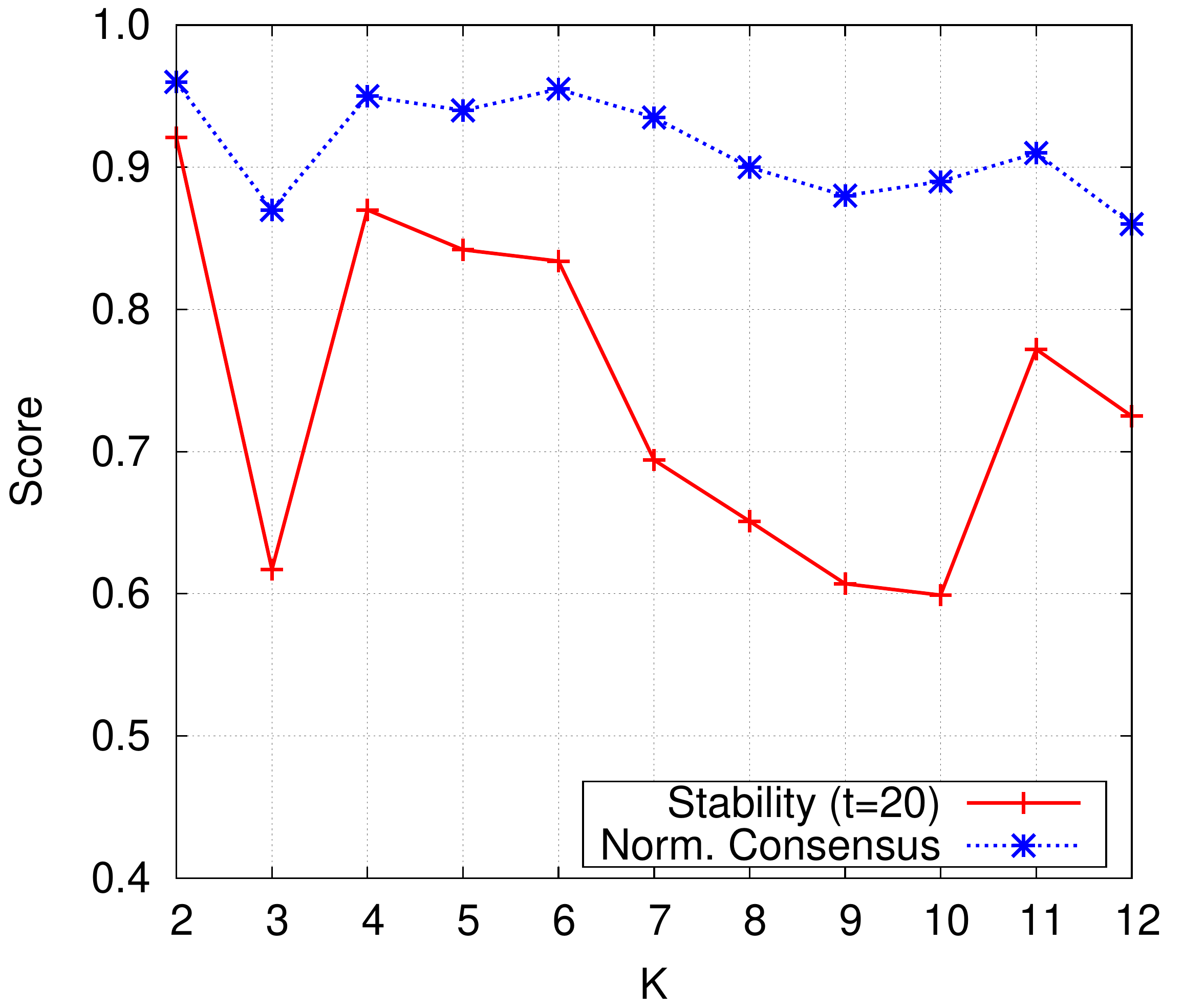}}
\hskip 0.8em
\subfigure[\emph{wikipedia-low}]{\includegraphics[width=0.48\linewidth]{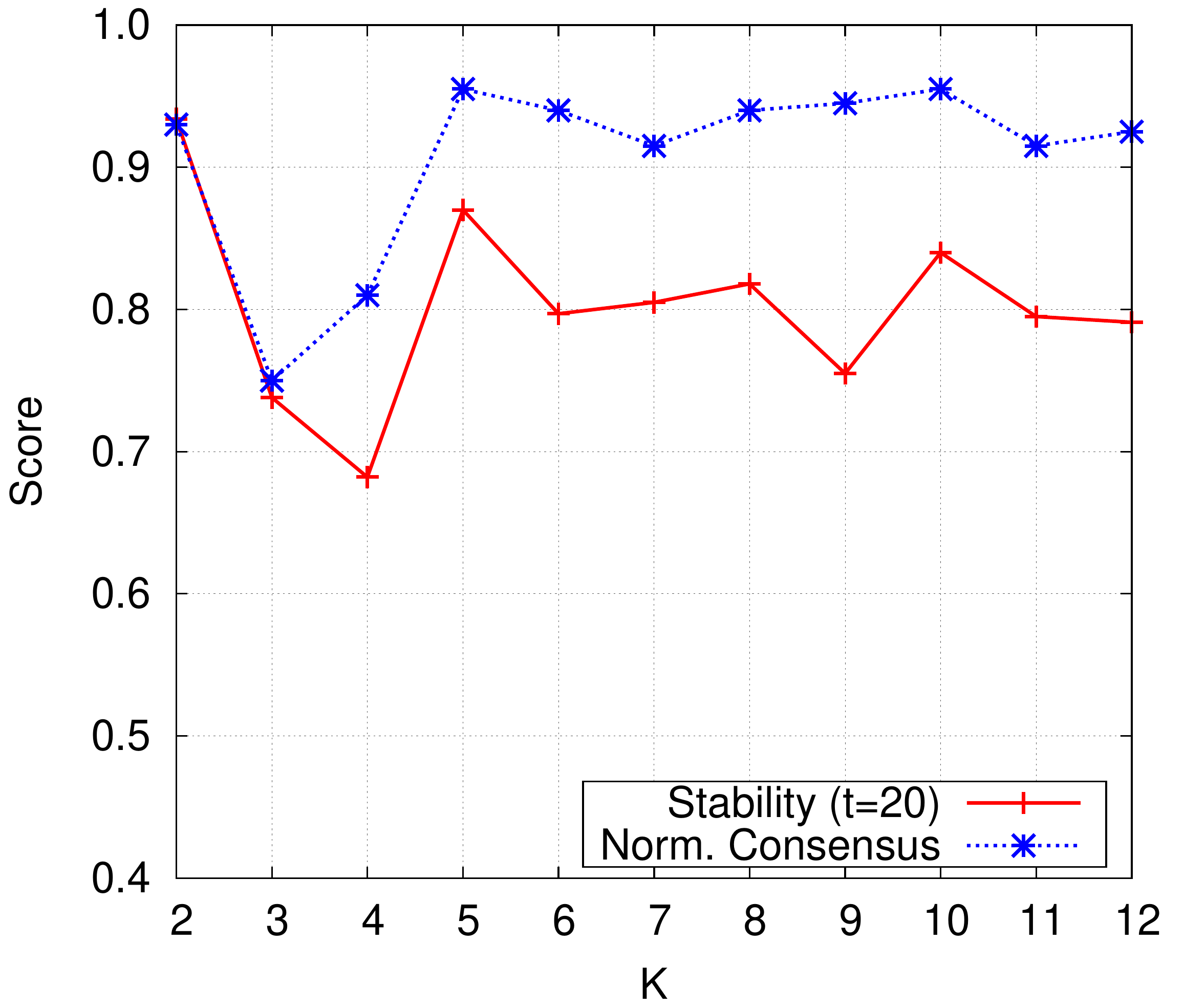}}
\vskip -1em
\caption{Comparison of plots generated for stability analysis ($t=20$) and consensus matrix analysis for values of $k \in [2,12]$.}
\label{fig:results2}
\end{figure}

Figures \ref{fig:results1} and \ref{fig:results2} show plots generated on the eight corpora for $k \in [2,12]$, comparing the proposed stability method with the consensus method from \cite{brunet04metagenes}. Although both measures can produce values in the range $[0,1]$, in all experiments the observed consensus scores were $> 0.8$ and often close to 1. Therefore, for the purpose of plotting the results, we apply min-max normalization to the consensus scores (with minimum value 0.8) to rescale the values to a more interpretable range.

We now summarize the results for each of the corpora in detail. The \emph{bbc} corpus contains five well-separated annotated categories for news articles, such as ``business'' and ``entertainment''. Therefore it is unsurprising that in \reffig{fig:results1}(a) we  find a strong peak for both methods at $k=5$, with a sharp fall-off for the stability method after this point. This reflects the fact that the five categories are accurately recovered by NMF. For the \emph{bbcsport} corpus, which also has five ground truth news categories, we see a peak at $k=4$, followed by a lower peak at $k=5$  -- see \reffig{fig:results1}(b). The consensus method also exhibits a peak at this point. Examining the top terms for the reference ranking set indicates that the two smallest categories, ``athletics'' and ``tennis'' have been assigned to a single larger topic, while the other three categories are clearly represented as topics.
 
In the ground truth for the \emph{guardian-2013} corpus, each article is labeled based upon the section in which it appeared on the {\tt{guardian.co.uk}} website. From \reffig{fig:results1}(c) we see that the stability method correctly identifies a peak at $k=6$ corresponding to the six sections in the corpus, which is not found by the consensus method. However, both methods also suggest a more coarse clustering at $k=3$. Inspecting the reference ranking set (see \reftab{tab:topics}(a)) suggests an intuitive explanation -- ``books'', ``fashion'' and ``music'' sections were merged in a single culture-related topic, documents labeled as ``politics'' and ``business'' were clustered together, while ``football'' remains as a distinct topic.

Articles in the \emph{irishtimes-2013} corpus also have annotated labels based on their publication section on {\tt{irishtimes.com}}.
In \reffig{fig:results1}(d) we see high scores at $k=2$ for both methods, and a subsequent peak identified by the stability method at $k=7$, corresponding to the seven publication sections. In the former case, the top ranked reference set terms indicate a topic related to sports and a catch-all news topic -- see \reftab{tab:topics}(b).

Next we consider the two corpora of news articles coming from the New York Times Annotated Corpus. Interestingly, for \emph{nytimes-1999}, in \reffig{fig:results2}(a) both methods exhibit a trough for $k=4$ topics, even though the ground truth for this corpus contains four news article categories. Inspecting the term rankings shown in\reftab{tab:topics}(c) provide a potential explanation of this instability: across the 100 factorization results, the ground truth ``sports'' category is often but not always split into two topics relating to baseball and basketball. 
%
For the \emph{nytimes-2003} corpus, which contains seven article categories, both methods produce high scores at $k=2$, with subsequent peaks at $k=4$ and $k=7$ -- see \reffig{fig:results2}(b). As with the \emph{irishtimes-2013} corpus, the highest-level structures indicate a simple separation between sports articles and other news. The reference topics at $k=4$ indicates that smaller categories among the New York Times articles, such as ``automobiles'' and ``dining \& wine'' do not appear as strong themes in the data.

Finally, we consider the two collections of Wikipedia pages, where pages are given labels based on their assignment to WikiProjects\footnote{See \url{http://en.wikipedia.org/wiki/Wikipedia:WikiProject}} at varying levels of granularity.
For \emph{wikipedia-high}, from \reffig{fig:results2}(c) we see that both methods achieve high scores for $k=2$ and $k=4$ topics. In the case of the former, the top terms in the reference ranking set indicate a split between Wikipedia pages related to music and all other pages (\reftab{tab:topics}(d)). While at $k=4$ (\reftab{tab:topics}(e)), we see coherent topics covering ``music'', ``sports'', ``space'', and a combination of the ``military'' \& ``transportation'' WikiProject labels. The ``medicine'' WikiProject is not clearly represented as a topic at this level.
In the case of \emph{wikipedia-low}, which contains ten low-level page categories, both methods show spikes at $k=5$, and $k=10$. At $k=5$, NMF recovers topics related to ``ice hockey'', ``cricket'', ``World War I'', a topic covering a mixture of musical genres, and a seemingly incoherent group that includes all other pages. The relatively high stability score achieved at this level (0.87) suggests that this configuration regularly appeared across the 100 NMF runs.

\begin{table}[!t]
\centering
\caption{Examples of top 10 terms for reference ranking sets generated by NMF on a number of text corpora for different values of $k$.}
\label{tab:topics}
\scriptsize{
\subtable[\emph{guardian-2013} ($k=3$)]
{
\begin{tabular}{|c|c|c|c|}
\hline
\;\emph{Rank}\; & \;\;\emph{Topic 1}\;\; & \;\;\emph{Topic 2}\;\; & \;\;\emph{Topic 3}\;\;\\ \hline
\;1 & book & league & bank\\ 
\;2 & music & club & government\\ 
\;3 & fashion & season & labour\\ 
\;4 & people & team & growth\\ 
\;5 & life & players & uk\\ 
\;6 & album & united & economy\\ 
\;7 & time & manager & tax\\ 
\;8 & novel & game & company\\ 
\;9 & love & football & party\\ 
\;10 & world & goal & market\\ \hline
\end{tabular}
}
\hskip 1.5em
\subtable[\emph{irishtimes-2013} ($k=2$)]
{
\begin{tabular}{|c|c|c|}
\hline
\;\emph{Rank}\; & \;\;\emph{Topic 1}\;\; & \;\;\emph{Topic 2}\;\;\\  \hline
\;1 & game & cent\\ 
\;2 & against & government\\ 
\;3 & team & court\\ 
\;4 & ireland & health\\ 
\;5 & players & ireland\\ 
\;6 & time & minister\\ 
\;7 & cup & people\\ 
\;8 & back & tax\\ 
\;9 & violates & dublin\\ 
\;10 & win & irish\\   \hline
\end{tabular}
}
\vskip 0.75em
\subtable[\emph{nytimes-1999} ($k=4$)]
{
\begin{tabular}{|c|c|c|c|c|}
\hline
\;\emph{Rank}\; & \;\;\emph{Topic 1}\;\; & \;\;\emph{Topic 2}\;\; & \;\;\emph{Topic 3}\;\; & \;\;\emph{Topic 4}\;\;\\ \hline
\;1 & game & company & yr & mets\\ 
\;2 & knicks & stock & bills & yankees\\ 
\;3 & team & market & bond & game\\ 
\;4 & season & business & rate & inning\\ 
\;5 & coach & companies & infl & valentine\\ 
\;6 & points & shares & bds & season\\ 
\;7 & play & stocks & bd & torre\\ 
\;8 & league & york & month & baseball\\ 
\;9 & players & investors & municipal & run\\ 
\;10 & sprewell & bank & buyer & clemens\\ \hline
\end{tabular}

}
\hskip 1.5em
\subtable[\emph{wikipedia-high} ($k=2$)]
{
\begin{tabular}{|c|c|c|}
\hline
\;\emph{Rank}\; & \;\;\emph{Topic 1}\;\; & \;\;\emph{Topic 2}\;\;\\  \hline
\;1 & album & team\\ 
\;2 & band & war\\ 
\;3 & song & star\\ 
\;4 & music & air\\ 
\;5 & released & season\\ 
\;6 & songs & aircraft\\ 
\;7 & chart & ship\\ 
\;8 & video & army\\ 
\;9 & rock & line\\ 
\;10 & albums & world\\  \hline
\end{tabular}
}
\vskip 0.75em
\subtable[\emph{wikipedia-high} ($k=4$)]
{
\begin{tabular}{|c|c|c|c|c|}
\hline
\;\emph{Rank}\; & \;\;\emph{Topic 1}\;\; & \;\;\emph{Topic 2}\;\; & \;\;\emph{Topic 3}\;\; & \;\;\emph{Topic 4}\;\;\\ \hline
\;1 & album & war & team & star\\ 
\;2 & band & air & season & planet\\ 
\;3 & song & ship & race & sun\\ 
\;4 & music & aircraft & league & earth\\ 
\;5 & released & army & game & stars\\ 
\;6 & songs & ships & championships & orbit\\ 
\;7 & chart & squadron & games & mass\\ 
\;8 & video & battle & cup & planets\\ 
\;9 & rock & station & world & system\\ 
\;10 & albums & british & championship & solar\\  \hline
\end{tabular}
}
\vskip 0.75em
\subtable[\emph{wikipedia-low} ($k=5$)]
{
\begin{tabular}{|c|c|c|c|c|c|}
\hline
\;\emph{Rank}\; & \;\;\emph{Topic 1}\;\; & \;\;\emph{Topic 2}\;\; & \;\;\emph{Topic 3}\;\; & \;\;\emph{Topic 4}\;\; & \;\;\emph{Topic 5}\;\;\\  \hline
\;1 & season & album & cricket & division & opera\\ 
\;2 & league & band & test & infantry & stakes\\ 
\;3 & team & released & match & battalion & race\\ 
\;4 & nhl & metal & innings & war & car\\ 
\;5 & hockey & music & runs & battle & racing\\ 
\;6 & games & song & wickets & brigade & engine\\ 
\;7 & cup & tour & against & army & old\\ 
\;8 & game & jazz & australia & regiment & horse\\ 
\;9 & goals & songs & england & german & stud\\ 
\;10 & club & albums & wicket & squadron & derby\\ \hline
\end{tabular}
}
}
\end{table}

\subsection{Discussion}
Overall, it is interesting to observe that, for a number of data sets, both methods evaluated here exhibited peaks at $k=2$, where one might expect far more fine-grained topics in these types of data sets. This results from high agreement between the term ranking sets generated at this level of granularity.
A closer inspection of document membership weights for these cases shows that this phenomenon generally arises from the repeated appearance of one small ``outlier'' topic and one large ``merged'' topic encompassing the rest of the documents in the corpus  (\eg the examples shown in \reftab{tab:topics}(b,d)). In a few cases we also see that the ground truth  does not always correspond well to the actual data (\eg for the sports-related articles in \emph{nytimes-1999}). This problem arises from time to time when meta-data is used to provide a ground truth in machine learning benchmarking experiments \cite{lee13community}.

In relation to computational time, the requirement to run a complete hierarchical clustering on the document-document consensus matrix before calculating cophenetic correlations leads to substantially longer running times on all corpora, when compared to the stability analysis method using a reference ranking set. In addition, the latter can be readily parallelized, as agreement scores can be calculated independently for each of the factorization results. 
\section{Conclusion}
\label{sec:conc}

A key challenge when applying topic modeling is the selection of an appropriate number of topics $k$. We have proposed a new method for choosing this parameter using a term-centric stability analysis strategy, where a higher level of agreement between the top-ranked terms for topics generated across different samples of the same corpus indicates a more suitable choice. Evaluations on a range of text corpora have suggested that this method can provide a useful guide for selecting one or more values for $k$. 

While our experiments have focused on the application of the proposed method in conjunction with NMF, the use of term rankings rather than raw factor values or probabilities means that it can potentially generalize to any topic modeling approach that can represent topics as ranked lists of terms. This includes probabilistic techniques such as LDA, together with more conventional partitional algorithms for document clustering such as $k$-means and its variants. In further work, we plan to examine the usefulness of stability analysis in conjunction with alternative algorithms.

\vspace{3 mm}\noindent{\bf{Acknowledgements.}}
This publication has emanated from research conducted with the financial support of Science Foundation Ireland (SFI) under Grant Number SFI/12/RC/2289.


\bibliographystyle{splncs03}
\bibliography{topics_arxiv}

\begin{thebibliography}{10}
\providecommand{\url}[1]{\texttt{#1}}
\providecommand{\urlprefix}{URL }

\bibitem{arora12beyond}
Arora, S., Ge, R., Moitra, A.: {Learning topic models -- Going beyond {SVD}}.
  In: Proc. 53rd Symp. Foundations of Computer Science. pp. 1--10. IEEE (2012)

\bibitem{bae06coala}
Bae, E., Bailey, J.: Coala: A novel approach for the extraction of an alternate
  clustering of high quality and high dissimilarity. In: Proc. 6th
  International Conference on Data Mining. pp. 53--62. IEEE (2006)

\bibitem{bendavid07stability}
Ben-David, S., P{\'a}l, D., Simon, H.U.: Stability of k-means clustering. In:
  Learning Theory, pp. 20--34. Springer (2007)

\bibitem{bertoni05stability}
Bertoni, A., Valentini, G.: Random projections for assessing gene expression
  cluster stability. In: Proc. IEEE International Joint Conference on Neural
  Networks (IJCNN'05). vol.~1, pp. 149--154 (2005)

\bibitem{blei03lda}
Blei, D.M., Ng, A.Y., Jordan, M.I.: Latent dirichlet allocation. Journal of
  Machine Learning Research  3,  993--1022 (2003)

\bibitem{bout08headstart}
Boutsidis, C., Gallopoulos, E.: {SVD based initialization: A head start for
  non-negative matrix factorization}. Pattern Recognition  (2008)

\bibitem{brunet04metagenes}
Brunet, J.P., Tamayo, P., Golub, T.R., Mesirov, J.P.: Metagenes and molecular
  pattern discovery using matrix factorization. Proc. National Academy of
  Sciences  101(12),  4164--4169 (2004)

\bibitem{dewall08stability}
De~Waal, A., Barnard, E.: Evaluating topic models with stability. In: 19th
  Annual Symposium of the Pattern Recognition Association of South Africa
  (2008)

\bibitem{fagin03topk}
Fagin, R., Kumar, R., Sivakumar, D.: Comparing top $k$ lists. SIAM Journal on
  Discrete Mathematics  17(1),  134--160 (2003)

\bibitem{greene05pkdd}
Greene, D., Cunningham, P.: Producing accurate interpretable clusters from
  high-dimensional data. In: Proc. 9th European Conference on Principles and
  Practice of Knowledge Discovery in Databases (PKDD'05). pp. 486--494 (2005)

\bibitem{hofmann99probabilistic}
Hofmann, T.: Probabilistic latent semantic analysis. In: Proc. 15th Conference
  on Uncertainty in Artificial Intelligence. pp. 289--296. Morgan Kaufmann
  (1999)

\bibitem{hutchins08position}
Hutchins, L.N., Murphy, S.M., Singh, P., Graber, J.H.: Position-dependent motif
  characterization using non-negative matrix factorization. Bioinformatics
  24(23),  2684--2690 (2008)

\bibitem{jaccard12index}
Jaccard, P.: The distribution of flora in the alpine zone. New Phytologist
  11(2),  37--50 (1912)

\bibitem{kendall90rank}
Kendall, M., Gibbons, J.D.: Rank Correlation Methods. Edward Arnold, London
  (1990)

\bibitem{kuhn55hungarian}
Kuhn, H.W.: The hungarian method for the assignment problem. Naval Research
  Logistics Quaterly  2,  83--97 (1955)

\bibitem{kumar10rankings}
Kumar, R., Vassilvitskii, S.: Generalized distances between rankings. In: Proc.
  19th International Conference on World Wide Web. pp. 571--580. ACM (2010)

\bibitem{lange04article}
Lange, T., Roth, V., Braun, M.L., Buhmann, J.M.: Stability-based validation of
  clustering solutions. Neural Computation  16(6),  1299--1323 (2004)

\bibitem{law03boot}
Law, M., Jain, A.K.: Cluster validity by bootstrapping partitions. Tech. Rep.
  MSU-CSE-03-5, University of Washington (February 2003)

\bibitem{lee13community}
Lee, C., Cunningham, P.: Community detection: effective evaluation on large
  social networks. Journal of Complex Networks  (2013)

\bibitem{lee99nmf}
Lee, D.D., Seung, H.S.: Learning the parts of objects by non-negative matrix
  factorization. Nature  401,  788--91 (October 1999)

\bibitem{levine01resampling}
Levine, E., Domany, E.: Resampling method for unsupervised estimation of
  cluster validity. Neural Computation  13(11),  2573--2593 (2001)

\bibitem{lin07gradient}
Lin, C.: {Projected gradient methods for non-negative matrix factorization}.
  Neural Computation  19(10),  2756--2779 (2007)

\bibitem{newman10aet}
Newman, D., Lau, J.H., Grieser, K., Baldwin, T.: {Automatic Evaluation of Topic
  Coherence}. In: Proc. Conf. North American Chapter of the Association for
  Computational Linguistics (HLT '10). pp. 100--108 (2010)

\bibitem{sandhaus08nytimes}
Sandhaus, E.: {The New York Times Annotated Corpus}. Linguistic Data
  Consortium, Philadelphia  6(12),  e26752 (2008)

\bibitem{steyvers07topicmodels}
Steyvers, M., Griffiths, T.: Probabilistic topic models. Handbook of latent
  semantic analysis  427(7),  424--440 (2007)

\bibitem{wang12group}
Wang, Q., Cao, Z., Xu, J., Li, H.: Group matrix factorization for scalable
  topic modeling. In: Proc. 35th SIGIR conference on Research and development
  in Information Retrieval. pp. 375--384. ACM (2012)

\bibitem{webber10indefinite}
Webber, W., Moffat, A., Zobel, J.: A similarity measure for indefinite
  rankings. ACM Transactions on Information Systems (TOIS)  28(4), ~20 (2010)

\end{thebibliography}

\end{document}